%% file: root.tex
\documentclass[letterpaper, 10 pt, conference]{ieeeconf}  

\IEEEoverridecommandlockouts    

\usepackage{graphicx} 
\usepackage{color} 
\usepackage[hidelinks]{hyperref}
\usepackage[font=footnotesize]{caption}
\usepackage{subcaption}
\usepackage{bm}
\usepackage{amsmath} 
\usepackage{amssymb}  
\usepackage{threeparttable}
\usepackage{booktabs} 
\usepackage{multirow}
\usepackage[ruled,linesnumbered]{algorithm2e}
\newcommand{\trsp}{\mathsf{T}}
\newcommand{\etal}{\MakeLowercase{\textit{et al. }}}

\title{\LARGE \bf A Laser-based Dual-arm System for Precise Control\\ of Collaborative Robots}

\author{Jo\~ao Silv\'erio, Guillaume Clivaz and Sylvain Calinon
\thanks{The authors are with the Idiap Research Institute, Martigny, Switzerland (\texttt{name.surname@idiap.ch}).}\thanks{This work was supported by Innosuisse (COB'HOOK project).} \thanks{We would like to thank Mr Teguh Santoso Lembono and Dr Yanlong Huang for their feedback on the manuscript.}}

\begin{document}

\maketitle
\thispagestyle{empty}
\pagestyle{empty}

\begin{abstract}
Collaborative robots offer increased interaction capabilities at relatively low cost but in contrast to their industrial counterparts they inevitably lack precision. Moreover, in addition to the robots' own imperfect models, day-to-day operations entail various sources of errors that despite being small rapidly accumulate. This happens as tasks change and robots are re-programmed, often requiring time-consuming calibrations. These aspects strongly limit the application of collaborative robots in tasks demanding high precision (e.g. watch-making).
We address this problem by relying on a dual-arm system with laser-based sensing to measure relative poses between objects of interest and compensate for pose errors coming from robot proprioception. Our approach leverages previous knowledge of object 3D models in combination with point cloud registration to efficiently extract relevant poses and compute corrective trajectories. This results in high-precision assembly behaviors. The approach is validated in a needle threading experiment, with a $150\mu m$ thread and a $300\mu m$ hole, and a USB insertion task using two 7-axis Panda robots.
\end{abstract}

\input{sections/introduction.tex}
\input{sections/relatedwork.tex}
\input{sections/proposedapproach.tex}
\input{sections/experiments.tex}
\input{sections/discussion.tex}
\input{sections/conclusion.tex}


\bibliography{root.bbl}

\end{document}

%% file: sections/introduction.tex
\section{INTRODUCTION}

\begin{figure}
	\centering
	\includegraphics[width=1.0\columnwidth]{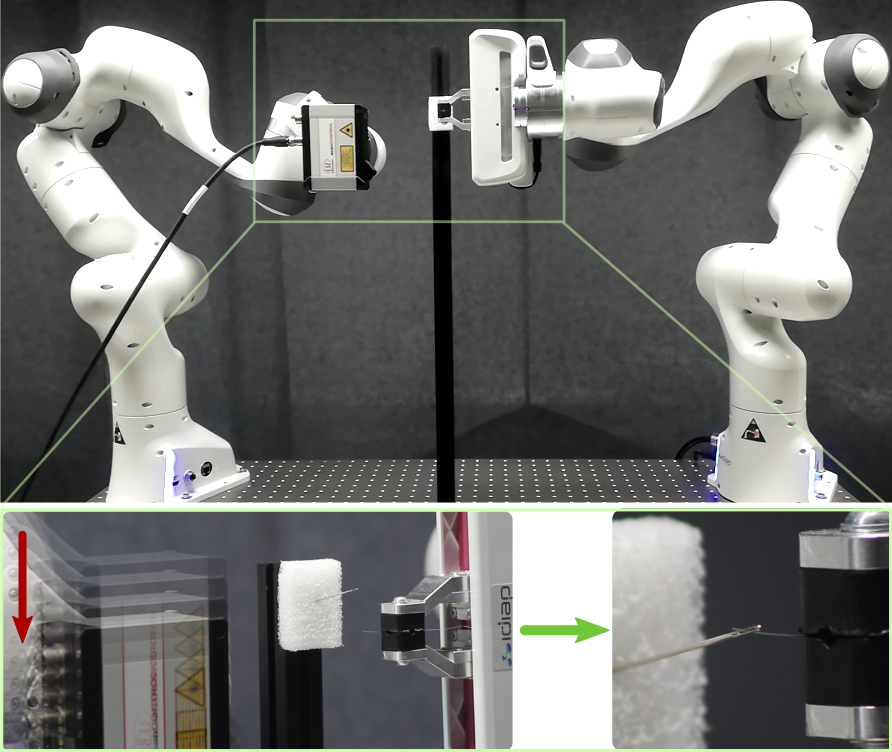}
	\caption{Proposed dual-arm system for high-precision insertion tasks performing a needle threading operation. The \textit{sensing} arm (left) moves a high-resolution sensing device (in this case a laser scanner) in the direction of the red arrow to compute the insertion poses with high precision. The \textit{assembling} arm (right) performs the insertion by tracking a low amplitude relative trajectory which ensures the insertion, overcoming existing modeling errors.}
	\label{fig:problem}
\end{figure}
\begin{figure}
	\centering
	\includegraphics[width=1.0\columnwidth]{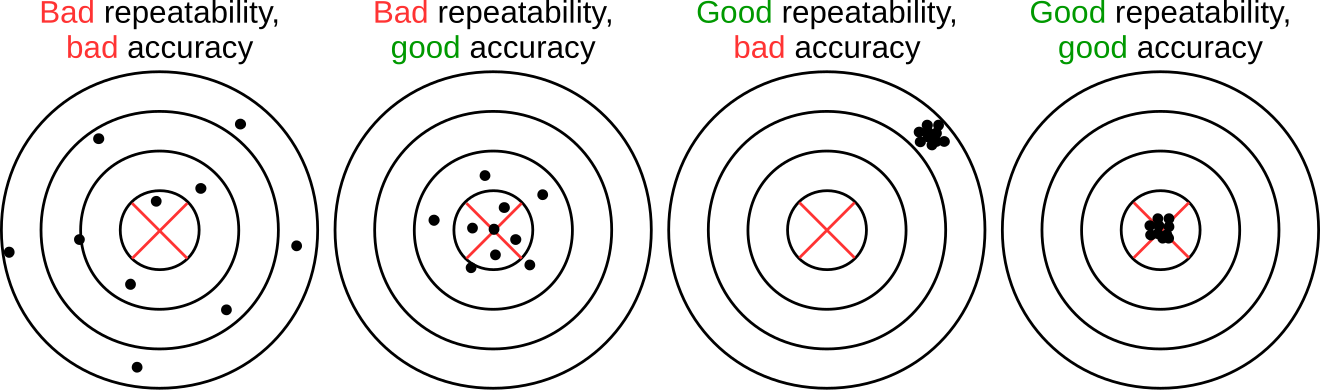}
	\caption{Difference between repeatability and accuracy as per ISO norm 9283:1998 \cite{ISO9283}. Each black point corresponds to an illustrative robot motion, directed to the red `$\times$' in the middle. Modern collaborative robots tend to belong to the third case (better repeatability than accuracy). Image reproduced from \cite{blogrobotiq}.}
	\label{fig:acc_rep}
\end{figure}

In recent years, small parts assembly has become a popular research topic in robotics \cite{Chatzilygeroudis20, Kimble20}. With the advent of collaborative robots, such as the Franka Emika Panda robots in Fig.~\ref{fig:problem}, this research direction gains special relevance. As safe, compliant and affordable platforms, collaborative robots often rely on torque control and less costly designs when compared to their industrial counterparts. This comes at the expense of precision, which is especially noticeable in tasks where tolerance is in the sub-millimeter range, such as needle threading (Fig.~\ref{fig:problem}). Indeed, despite having good \textit{repeatability}, the \textit{accuracy} of most collaborative robots does not allow them to rely on proprioception alone for this kind of tasks (Fig.~\ref{fig:acc_rep}). With the dynamic nature of modern factories, in which collaborative robots are meant to excel, and all the associated perception and re-programming, it becomes clear that robust solutions for manipulating small parts are required.

A popular approach to circumvent imperfect proprioception is to combine control with sensory feedback. This type of approach takes inspiration from how humans, who have centimeter-level repeatability, perform manipulation tasks by incorporating feedback from vision \cite{Chaumette06} or touch \cite{Pastor11}.
In high-precision manipulation tasks, tactile feedback is often poor (e.g.~needle threading), undesirable (e.g.~small components are often fragile) or unreliable (due to low magnitude of the forces involved). Hence, visual or spatial perception are the most promising types of feedback. Here, we propose a dual-arm solution with laser scanning to address the precision problem (Section \ref{sec:approach}). The general idea is to have one arm responsible for manipulation (e.g.~assembly, insertion) while the other measures the poses between the object being manipulated and the part where it will be placed (e.g.~thread tip and needle hole in Fig.~\ref{fig:problem}). Having the relevant poses, a low-amplitude \textit{relative trajectory} that compensates pose errors is computed and applied to the manipulator such that the object pose is corrected, resulting in high-precision assembly behaviors. Our approach leverages previous knowledge of the objects' 3D models in combination with registration of high-resolution point clouds (we use a laser scanner with a reference $1.5\mu m$ depth resolution and 2048 points/profile along the $x$-axis) to efficiently extract relevant poses. For this aspect, we argue that exploiting previously available CAD models, similarly to \cite{Kimble20, Tamadazte10, Roveda20}, can have a strong practical impact in high-precision manipulation for collaborative robots.
In summary, this paper advances the state-of-the-art in four main ways by:
\begin{enumerate}
	\item introducing a novel approach to perform sub-millimeter insertion tasks using collaborative robots;\label{point_1}
	\item relying on laser scanning in the $\mu$-meter range;
	\item being a dual-arm approach, which provides the possibility to implement rich sensing behaviors; and
	\item leveraging pre-existing CAD models for sub-millimeter assembly tasks.
\end{enumerate}

To the best of our knowledge, no previous work has addressed the issue of sub-millimeter manipulation with collaborative robots using high-resolution laser sensing. The impact of our solution is not limited to the applications showcased in Sections \ref{sec:exp}--\ref{sec:exp_usb}, but has potential in any area requiring small part manipulation, e.g.~watch-making \cite{Chatzilygeroudis20} and surgery \cite{Staub10} (we discuss possible limitations in Section \ref{sec:discussion}).

%% file: sections/relatedwork.tex
\section{RELATED WORK}

Visual servoing \cite{Chaumette06} is a classical approach for handling uncertain feedback signals, whether proprioceptive or from the environment. Therefore, it has been a popular choice in past works on small parts manipulation \cite{Staub10,Tamadazte10,Tamadazte11,Xing14,Ma19}. The answer to what type of visual servoing should be used often boils down to the specificities of the application such as the available hardware and the properties of the scene and objects involved (e.g.~lighting, textures). Recent work from Roveda \etal on 6D pose estimation \cite{Roveda20} brings forward the insight that CAD-based methods are well suited for dynamic scenarios, such as Industry-4.0-type plants where appropriate lighting is often not guaranteed and where objects often have little texture. 
Similarly to our work, \cite{Roveda20} relies on point cloud registration to find object poses, however at a much larger object scale. 
We follow their insight and rely on CAD models to obtain reference point clouds for registration. When not available, we use the \textit{sensing arm} to model relevant objects. Interestingly, Tamadazte \etal \cite{Tamadazte10} also proposed a CAD-model-based tracking method, but applied to visual servoing (without point cloud registration). Their micro-assembly approach is, however, not trivial to extend to dynamic environments like the ones we envision. 

Our approach can be seen as a form of visual servoing where the sensory feedback is not provided continuously but every time a new scan is performed. Due to the low amplitude of the corrective trajectory that is computed, proprioception errors do not accumulate to a degree where the task is compromised. Additionally, the fact that the computed trajectory is relative, i.e., computed between two object poses represented in the same reference frame, alleviates the need for very precise calibrations. The distinguishing feature of our work on the sensing side is the very precise point clouds that we operate on, which come from the use of a high-resolution laser line scanner \cite{microepsilon}. Depth sensing is a popular approach to modeling objects of various sizes, e.g.~bridges \cite{Paul11} or daily-life objects \cite{Krainin11}. We here study the applicability at much lower scales.

Finally, we highlight the high reproducibility of this work. By using popular collaborative robots, commercially available sensors, benchmark tools \cite{Kimble20} and common daily-life objects (e.g.~needles, threads), our results are easy to replicate when compared to other works on the topic that require more specialized robots, objects and setups, e.g. \cite{Tamadazte10, Staub10, Ma19}.

%
%
%

%
%
%
%
%
%
%
%
%
%
%
%

%% file: sections/proposedapproach.tex
\section{PROPOSED APPROACH}
\label{sec:approach}

\subsection{Overview}

The proposed approach revolves around the setup in Fig.~\ref{fig:problem}. In that figure we see one robot holding a high-resolution laser scanner and another holding a thread. Throughout the paper we will refer to the former as the \textit{sensing} arm and the latter as the \textit{assembling} arm. While in Fig.~\ref{fig:problem} the assembling arm is holding a thread, generically this can be any object that needs to be inserted/fit into another object. We will assume that the action that the assembling arm must perform comes down to an insertion (threading, assembly or other).

\begin{figure}
	\centering
	\includegraphics[width=.9\columnwidth]{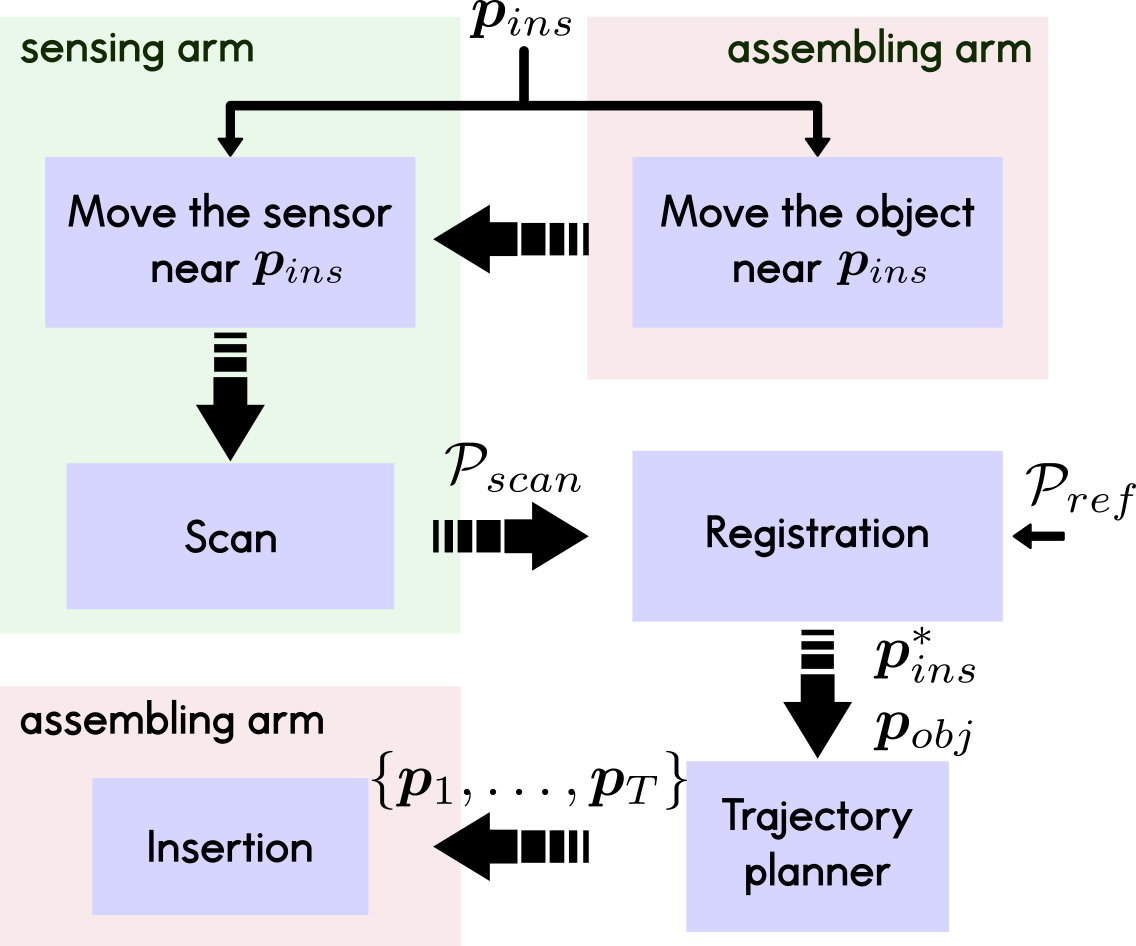}
	\caption{Diagram of the proposed dual-arm approach.}
	\label{fig:diagram}
\end{figure}

Figure \ref{fig:diagram} gives an overview of our method. Both arms start with the knowledge of the \textit{insertion pose}, ${\bm{p}_{ins} = \left[\bm{x}^\trsp_{ins}\> \bm{q}^\trsp_{ins} \right]^\trsp}$ where $\bm{x}_{ins}\in\mathbb{R}^3, \> \bm{q}_{ins}\in\mathcal{S}^3$. This pose represents an initial guess about where the insertion will take place in the robot workspace and is used to bring the laser scanner close enough that a point cloud of both the object to be inserted and its counterpart (where it will be inserted) can be collected. Both $\bm{p}_{ins}$ and the initial poses of both arms (see first two boxes in Fig.~\ref{fig:diagram}) can be collected beforehand, e.g.~by kinesthetic teaching.

Once the scanner is close enough to the insertion pose, a point cloud $\mathcal{P}_{scan}$ is collected. This point cloud can contain sub-point-clouds that correspond to both the object to be inserted and the insertion target (e.g.~a thread and a needle). With $\mathcal{P}_{scan}$, point cloud registration is performed (fourth box in Fig.~\ref{fig:diagram}), against a reference point cloud $\mathcal{P}_{ref}$, from which the poses of the object to be inserted $\bm{p}_{obj}$ and the insertion target $\bm{p}^*_{ins}$ are extracted. With $\bm{p}_{obj}$ and $\bm{p}^*_{ins}$, a trajectory that brings the end-effector to the final insertion pose is computed and sent to the assembling arm to be tracked (last two boxes in Fig.~\ref{fig:diagram}). Since the resulting trajectory will have a very low amplitude, any modeling errors (e.g. kinematic uncertainty) that might exist are unlikely to accumulate enough to compromise the insertion.

The approach that we propose is both modular and generic by design. For instance, any box in Fig.~\ref{fig:diagram} plays a distinct role that depends only on the completion of the previous one in the chain. Additionally, we impose no constraint on how each module is implemented, e.g.~the specific motion of the sensing arm or the registration algorithm. Next, we describe our proposed implementations and a few considerations.

\subsection{Scanning}

The sensing arm is responsible for obtaining a high-resolution point cloud $\mathcal{P}_{scan}$ of both the object to be inserted and the part where it will be inserted (both assumed rigid enough not to deform significantly during insertion). Note that there is the possibility to run multiple scans with different poses if necessary (e.g.~for improving $\mathcal{P}_{scan}$).  An important aspect to consider is the transformation between the sensor reference frame and the robot base. Here we used the technical drawing from the sensor manual to manually compute the pose of the scanner with respect to the robot base. Calibration techniques like the one introduced by Lembono \etal \cite{Lembono18} could be used either when an accurate model is unavailable or to refine an initial estimate. It should be noted, however, that the usual concerns with calibration precision are alleviated in our approach, since we rely on a \textit{relative trajectory}, given the poses  $\bm{p}_{obj}$, $\bm{p}^*_{ins}$ (Section \ref{sec:planning_insertion}).

\subsection{Point cloud registration}

From $\mathcal{P}_{scan}$ we estimate $\bm{p}_{obj}$ and $\bm{p}^*_{ins}$. For this we rely on point cloud registration. Note that, with a high-resolution laser scanner such as the one we use, the point density will be very high. This is crucial for high-precision insertions. Similarly to \cite{Roveda20} we pre-process $\mathcal{P}_{scan}$, as a means to remove noise  and keep only the most important points. 
We also need the original point clouds of the objects whose pose we intend to discover, which we refer to as $\mathcal{P}_{ref}$. One way to obtain $\mathcal{P}_{ref}$ is by scanning objects beforehand and storing their point clouds. However, as highlighted in \cite{Kimble20}, in many automation processes the CAD models of the objects are readily available. We propose to use them whenever possible to obtain $\mathcal{P}_{ref}$. A myriad of tools exist for converting CAD models to point clouds. Here we used Blender \cite{BOC18}.

\begin{algorithm}[h]
 \SetKwInOut{Input}{Input}
 \SetKwInOut{Output}{Output}
 \Input{Point clouds $\mathcal{P}_{scan}$, $\mathcal{P}_{ref}$; thresholds $\rho_{ICP}$, $\rho_{rot}$; initial estimate of object orientation $\bm{q}_{0}$}
 \Output{Pose of scanned object $\hat{\bm{p}}=\left[\hat{\bm{x}}^\trsp\> \hat{\bm{q}}^\trsp\right]^\trsp$}
 filter and downsample $\mathcal{P}_{scan}$\;
 initialize ICP best fitness score $f^*_{ICP} = 1e6$\;
 \While{$f^*_{ICP} > \rho_{ICP}$}
 {
  $\hat{\bm{p}}, \hat{\mathcal{P}} \leftarrow$ RANSAC($\mathcal{P}_{scan}$, $\mathcal{P}_{ref}$)\;
  \If{$\text{d}_\phi(\hat{\bm{q}}$,$\bm{q}_{0})$ $< \rho_{rot}$}
  {
   $f_{ICP},\bm{p}_{ICP} \leftarrow$ ICP($\mathcal{P}_{scan},\hat{\mathcal{P}}$)\;
	   \If{$f_{ICP} < f^*_{ICP}$}
	   {
	   	$f^*_{ICP} \leftarrow f_{ICP}$\;
	   	$\hat{\bm{p}} \leftarrow \bm{p}_{ICP}$\;
	   }
   }
 }
 \caption{Pose estimation via point cloud registration}
 \label{algo:pose_estimation}
\end{algorithm}

For pose estimation, we used Algorithm \ref{algo:pose_estimation}: a combination of RANSAC \cite{Fischler81}, for a coarse initial registration of $\mathcal{P}_{scan}$ and $\mathcal{P}_{ref}$, and ICP \cite{Besl92}, for refining the result. Particularly, we relied on their implementations from the Point Cloud Library (PCL) \cite{Rusu11}. From the end-effector pose, we compute an initial guess of the manipulated object orientation $\bm{q}_0$ that we use to filter out possible poor matches from RANSAC. For objects that are not attached to the robot, other heuristics can be used to compute $\bm{q}_0$ (e.g. a prior on pointing direction). When poor registration from RANSAC occurs, which can hinder ICP performance, it is most often a consequence of orientation mismatches. In order to quantify these, we compute an orientation error $\text{d}_\phi(\hat{\bm{q}},\bm{q}_0)$ \cite{Zeestraten17} between the RANSAC-estimated orientation $\hat{\bm{q}}$ and $\bm{q}_0$. We then evaluate it against a threshold $\rho_{rot}$ (set loosely enough to only reject unrealistic matches) to keep or reject the registration. Finally, we define a threshold $\rho_{ICP}$ to stop the algorithm when the distance between point clouds is small enough. Both thresholds are chosen empirically. In Algorithm \ref{algo:pose_estimation}, $f, \mathcal{P}, \bm{p}$ denote fitness scores, point clouds and poses, respectively. Only the outputs of RANSAC() and ICP() that are relevant to the approach are indicated in rows 4 and 6.


\subsection{Trajectory planning and insertion}
\label{sec:planning_insertion}
Using Algorithm \ref{algo:pose_estimation} we obtain the poses $\bm{p}^*_{ins}$ and $\bm{p}_{obj}$. After knowing these poses we use spline interpolation to compute a corrective Cartesian trajectory for the assembling arm $\{\bm{p}_1,\ldots,\bm{p}_T\}$, where $T$ is the time horizon, that compensates insertion pose errors. Since $\bm{p}^*_{ins}$ and $\bm{p}_{obj}$ are represented in the same reference frame (the one of the sensing arm), a trajectory that connects the two poses is \textit{relative} in nature, i.e.~it can be applied as an offset so that the robot moves incrementally from its current state. One positive side effect is that the need for a very precise calibration of the sensor (as well as between the robot's bases, which here was manually estimated) is considerably alleviated. In this work we used spline interpolation to plan relative trajectories but more elaborated planning can be applied.

%% file: sections/experiments.tex
\section{EXPERIMENT I - NEEDLE THREADING}
\label{sec:exp}

We applied the proposed approach to a needle threading experiment, a scenario that demands high precision.

\subsection{Setup}

The experimental setup is shown in Fig.~\ref{fig:problem}. The assembling arm holds a thread that is to be inserted in the needle hole. Our evaluation considers three different aspects: \textit{initial conditions}, \textit{testing settings} and \textit{threading strategies}. \textit{Initial conditions} refer to the robot configuration \textbf{before} it starts moving towards the needle. \textit{Testing settings} correspond to \textbf{different needle poses}.  \textit{Threading strategies} pertain to the robot configuration \textbf{when inserting} the thread and whether or not it uses sensory feedback. We now elaborate on these.

\paragraph{Initial conditions}
Every time the robot performs an insertion, regardless of the threading strategy and needle pose, it starts the motion at one of ten previously recorded joint configurations $\bm{q}^{(1)},\ldots,\bm{q}^{(10)}$. These were recorded such that the position of the end-effector at each joint configuration $\bm{q}^{(i)}$ is further away from the insertion pose than the previous, i.e. ${\|\bm{x}^{(i)}-\bm{x}_{ins}\| > \|\bm{x}^{(i-1)}-\bm{x}_{ins}\|, \>i=2,\ldots,10}$, where $\bm{x}^{(i)}=f(\bm{q}^{(i)})$ is the end-effector position and $f(.)$ is the forward kinematics function. 

\begin{figure}
	\centering
	\includegraphics[width=1.0\columnwidth]{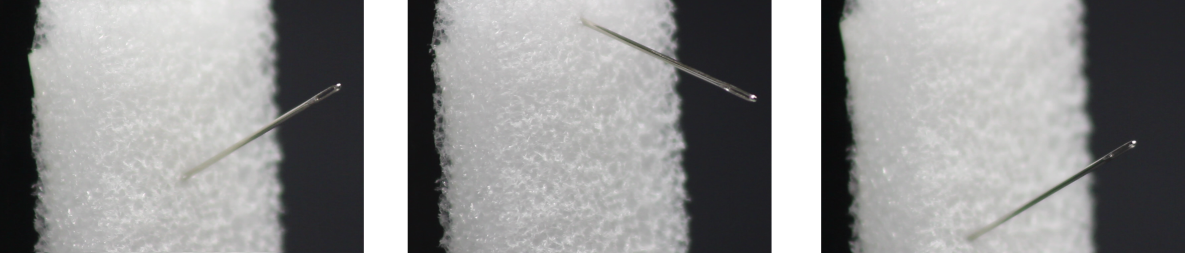}
	\caption{Three different needle poses used in the evaluation. We refer to each of these settings as \textit{needle 1}, \textit{needle 2} and \textit{needle 3}, by the order they appear in the figure.}
	\vspace{0.2cm}
	\label{fig:three_holes}
\end{figure}

\begin{figure}
	\centering
	\begin{subfigure}[b]{0.35\columnwidth}
		\centering
		\includegraphics[width=\textwidth]{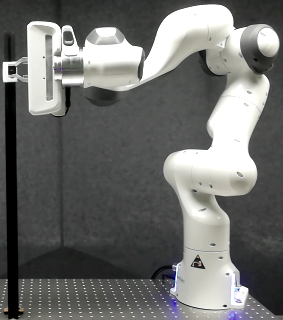}
		\caption{\footnotesize\centering Insertion configuration 1 (IC1).}
		\label{fig:pose1_proprio}
	\end{subfigure}
	\hspace{1cm}
	\begin{subfigure}[b]{0.35\columnwidth}
		\centering
		\includegraphics[width=\textwidth]{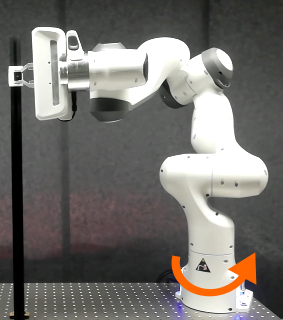}
		\caption{\footnotesize\centering Insertion  configuration 2 (IC2).}
		\label{fig:pose2_proprio}
	\end{subfigure}
	\vspace{0.5cm}
	\caption{Two different robot configurations used for threading. The effect of changing the robot configuration on the insertion performance was significant in the final results.}
	\label{fig:poses_proprio}
\end{figure}

\paragraph{Testing settings}
We report results for three different needle poses (Fig.~\ref{fig:three_holes}). Testing the approach with different needle poses allows us to evaluate its robustness when  task conditions change. At the scale of this task, variations such as the ones in Fig.~\ref{fig:poses_proprio}, which would normally seem negligible, can cause a strong impact on performance and are worth investigating. Prior to evaluations on any given needle pose, we performed an insertion on that needle and stored the pose of the assembling arm end-effector $\bm{p}_{ins}$. 
For each needle pose we compared three different threading strategies.

\paragraph{Threading strategies}

We categorize threading strategies by whether the robot uses the laser scanner (insertion with sensory feedback) or not (proprioception only). In the latter case, we further break down the evaluation into which of two possible joint configurations IC1 and IC2 (Fig. \ref{fig:poses_proprio}) the robot uses during insertion. The null space of the task was used to bias the inverse kinematics (IK) solution to stay close to IC1 or IC2. The difference between IC1 and IC2 is that the former is the configuration of the robot during the recorded insertion and the latter is the configuration obtained by taking IC1 and adding $\sim\pi/2$ to the first joint (the rest of the joints seen in Fig.\ref{fig:pose2_proprio} adapt accordingly so that the robot can accomplish the task). We hypothesize that IC1 will 
perform better than IC2 
since it 
requires smaller joint displacements with respect to when $\bm{p}_{ins}$ was recorded. In other words, IC2 will result in higher pose errors (more failed insertions). Specifically, the compared strategies were:

\begin{enumerate}
	\item \textbf{Proprioception only + IC1.} The robot threads the needle using proprioception only and IC1 (Fig.~\ref{fig:pose1_proprio}).
	\item \textbf{Proprioception only + IC2.} The robot threads the needle relying on proprioception and IC2 (Fig.~\ref{fig:pose2_proprio}).
	\item \textbf{Insertion with sensory feedback (laser scanner).} The robot threads the needle using IC2 
and compensates for insertion pose errors using sensory feedback as per Section \ref{sec:approach}.
\end{enumerate}

\subsection{Hardware and control}

In our experiments (including Section \ref{sec:exp_usb}) we used a Micro-Epsilon LLT3000-25/BL laser scanner \cite{microepsilon} attached to the left robot arm. We operated the scanner at its reference resolutions of $1.5\mu m$ (depth) and $~12\mu m$ ($x$-axis, 2048 points/profile). Moreover, for needle threading, we used a $150\mu m$-width thread and a needle with a  hole width in the range ${300-350\mu m}$  along the smallest axis. Notice both the low scale and tolerance of this insertion task.

As robot platform, we used two 7-DoF Franka Emika Panda arms. The repeatability of the Panda arm, as reported by the manufacturer, is $0.1mm$. To the best of our knowledge there is no official value for accuracy. In order to ensure the highest possible precision we used position control.

\begin{table}
	\vspace{0.5cm}
	\renewcommand{\arraystretch}{1.3} 
	\centering
	\caption{Success rates from needle threading experiment. Each number in the columns 'Needle 1 ... 3' corresponds to 10 insertion attempts.}
	\begin{threeparttable} 
		\begin{tabular}{p{2.5cm}cccc}
			\toprule[0.12em]
			\centering \multirow{2}{*}{\textbf{Strategy}} & \multicolumn{3}{c}{\textbf{Setting}} & \multirow{2}{*}{\textbf{Total}}\\
			\cmidrule{2-4}
			& Needle 1 & Needle 2 & Needle 3 &\\
			\midrule
			Proprioception only + IC1 & 50\% & 60\% & 70\% &\textbf{60\%}\\
			Proprioception only + IC2 & 0\% & 0\% & 0\% &\textbf{0\%}\\
			With laser scanner & 100\% & 100\%\tnote{$\dagger$} & 90\% &\textbf{96.7\%}\\
			\bottomrule[0.12em]
		\end{tabular}
	\begin{tablenotes}
		\item[$\dagger$] Includes 2 failures that were successful upon re-scanning.
	\end{tablenotes}
	\end{threeparttable}
	\label{tab:SR_needle}
	\vspace{-0.5cm}
\end{table}

\begin{figure*}
	\centering
	\includegraphics[width=0.98\textwidth]{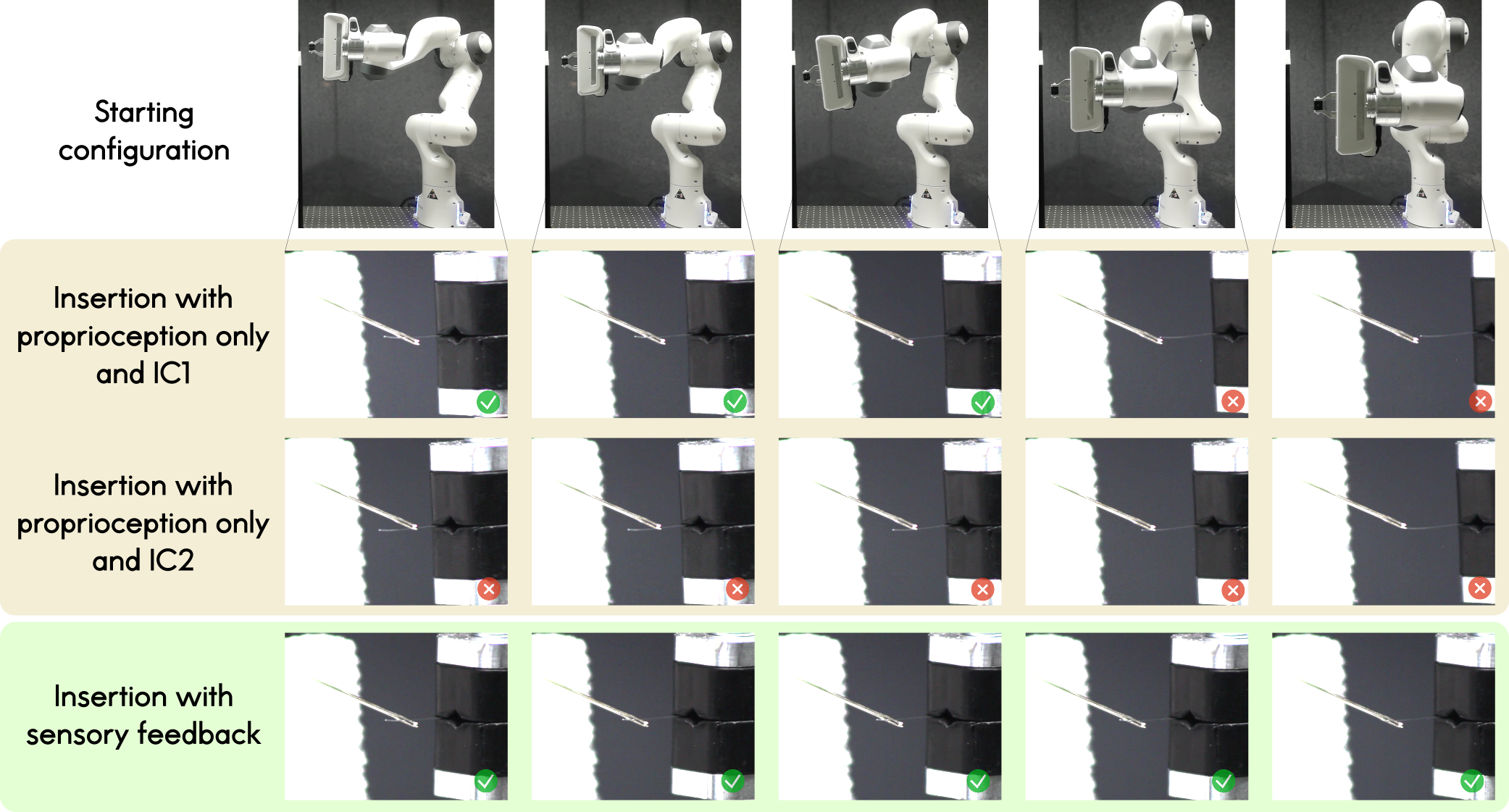}
	\caption{Snapshots of the needle threading experiment, where we show 5 out of 10 results for the second needle pose in Fig.~\ref{fig:three_holes}. The full results are reported in Table \ref{tab:SR_needle}. \textbf{Top row:} Initial poses from which the robot performed the insertion. \textbf{Second row:} Insertion with proprioception only and IC1 as robot configuration. \textbf{Third row:} Results for proprioception only and IC2. \textbf{Bottom row:} Insertion after correction from the laser scanner. Without sensory feedback, the performance deteriorates both with the distance to the insertion pose and with the insertion configuration. A detailed visualization including all cases is given in the supplementary videos.}
	\label{fig:threading_all}
\end{figure*}

\subsection{Results}

Here we describe the obtained results from each strategy. All the reported results (including those in Section \ref{sec:exp_usb}) can be seen in the supplementary videos, also available at \url{https://sites.google.com/view/laser-cobots/}.

\paragraph{Insertion with proprioception only and IC1}

Given $\bm{p}_{ins}$, the robot is commanded to reach that end-effector pose, starting from different poses in the workspace. Table \ref{tab:SR_needle} shows the obtained success rates for this strategy (first row). The results show that, on average, 
the robot managed to thread the needle 60\% of the times. Figure \ref{fig:threading_all} (second row) further shows this. As the initial poses are farther and farther away from the insertion pose, the success rate decreases. 
This can be due to a combination of the wider range of joint motions required to reach the insertion pose, 
uncertainties in the planned insertion trajectories (which have a bigger length as the robot starts farther away) and potentially the accumulation of small deformations of the (non-rigid) thread after several attempts. Nonetheless, owing to the $0.1mm$ repeatability, a 60\% success rate at this scale is noteworthy.

\paragraph{Insertion with proprioception only and IC2}

The results showed that threading is never successful when the insertion joint configuration differs from the original one by a non-negligible amount (see Table \ref{tab:SR_needle}, second row). The third row in Fig.~\ref{fig:threading_all} is clear in this aspect, with the thread always ending up either below or before the needle. The arguments given in the previous paragraph hold for this case as well. Here, since not only the starting poses, but also the insertion configuration differs by a significant amount, the previous observations are amplified.

\paragraph{Insertion with laser scanner}

\begin{figure}
	\centering
	\includegraphics[width=0.55\columnwidth]{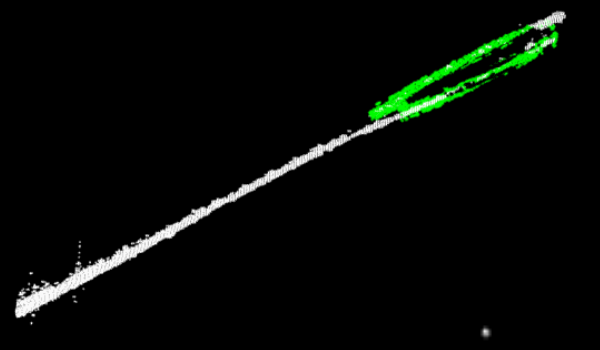}
	\caption{Example of point cloud obtained after scanning the needle and the thread. The diagonal cluster of points is the needle point cloud $\mathcal{P}_{scan}$. The registered model $\mathcal{P}_{ref}$ (green) after Algorithm \ref{algo:pose_estimation} is also shown. Note that in this experiment $\mathcal{P}_{ref}$ was obtained by scanning the needle a priori. The small point cloud on the bottom-right corner is the thread tip.}
	\label{fig:needle_point_cloud}
\end{figure}

Finally, using the proposed approach with laser scanning, the previous results were considerably improved. As seen in Table \ref{tab:SR_needle} insertion success rates reach values closer to 100\%.
In this scenario, the robot was commanded to thread the needle, by first using its proprioception alone as in the previous cases. Once it failed the insertion, the sensing arm performs a scan to obtain a point cloud of the needle and the thread. Using this point cloud it runs Algorithm \ref{algo:pose_estimation} to find the needle pose. Figure \ref{fig:needle_point_cloud} shows an example of a scan. Notice the high density of the cloud, for such a small object. The device resolution proves essential for this sort of tasks. In the bottom-right corner of the image we can see a small point cloud, which corresponds to the thread tip. In our experiments we selected the tip manually on the GUI, but this step can alternatively be automated, similarly to how the needle pose is found.
With the needle pose and the thread tip, the robot computes a low amplitude trajectory that compensates for the pose error, ensuring insertion. It is worth pointing out that the results in the third row of Table \ref{tab:SR_needle} include two attempts where the robot initially failed the insertion but, upon a re-scan, succeeded. Indeed, by bringing sensory feedback into the loop, one is able not only to detect errors but also to correct them. The bottom row of Fig.~\ref{fig:threading_all} shows 5 out of 10 successful insertions for the \textit{needle 2} setting. 

\section{EXPERIMENT II - USB CABLE INSERTION}
\label{sec:exp_usb}

\begin{figure}
	\centering
	\includegraphics[width=0.8\columnwidth]{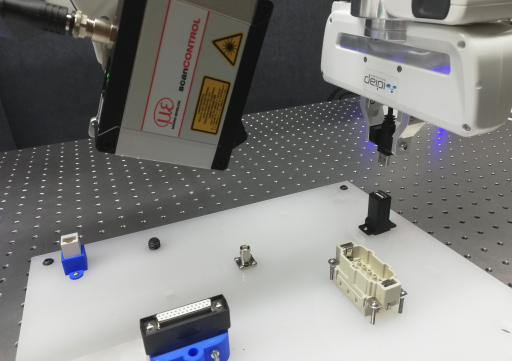}
	\\
	\vspace{0.1cm}
	\includegraphics[width=0.925\columnwidth]{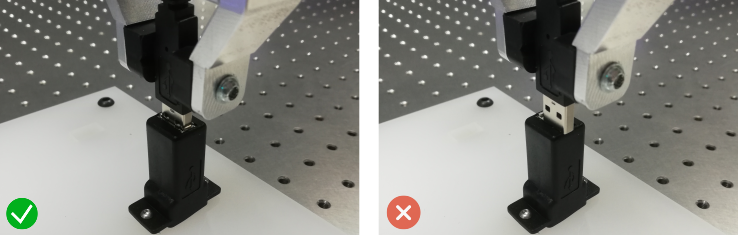}
	\caption{USB insertion setup with the NIST Task Board 1.}
	\label{fig:USB_setup}
\end{figure}

In a second experiment we used the NIST Task Board 1 \cite{Kimble20} to study the insertion of a USB plug (Fig.~\ref{fig:USB_setup}). This time we obtained both socket and plug point clouds from their CAD models, made available	 by NIST. Similarly to Section \ref{sec:exp}, we computed success rates for 10 insertions, where the robot started at 10 different pre-recorded configurations. Note that success rates are one of the performance metrics proposed in \cite{Kimble20}. We considered two different robot configurations IC1 and IC2 following the same convention as in Section \ref{sec:exp}. In this case, we evaluated the approach for one single socket pose. When inserting with sensory feedback, both socket and USB plug are scanned and registered in order to compute $\bm{p}_{obj}$ and $\bm{p}^*_{ins}$.

\begin{table}
	\vspace{0.5cm}
	\renewcommand{\arraystretch}{1.3} 
	\centering
	\caption{Success rates for 10 USB insertions.}
	\begin{tabular}{cc}
		\toprule[0.12em]
		\textbf{Strategy} & Total\\
		\midrule
		Proprioception only + IC1  & \textbf{90\%}\\
		Proprioception only + IC2 & \textbf{0\%}\\
		With laser scanner & \textbf{100\%}\\
		\bottomrule[0.12em]
	\end{tabular}
	\label{tab:SR_usb}
\end{table}

Table \ref{tab:SR_usb} shows the obtained success rates. As in the needle scenario, all evaluations in the first two rows consisted of reaching a pre-recorded successful pose $\bm{p}_{ins}$. 
Notice that, in the first row, the success rate is higher than the one obtained in the previous experiment. One possible explanation is that the  tolerance is slightly higher in this setup. Nevertheless, as in the previous case, when changing the joint configuration to IC2 the performance rapidly deteriorates (second row in Table \ref{tab:SR_usb}). 
In most practical applications, this effect will appear naturally since the robot needs to modify its configuration when object poses change. 
Similarly to Section \ref{sec:exp}, sensory feedback strongly improved the performance (last row of Table \ref{tab:SR_usb}). Figure \ref{fig:USB_pc} shows typical point clouds obtained for the USB plug. The high resolution once again stands out. The registration results were particularly robust, which is well-attested by the 100\% success rate.

%% file: sections/discussion.tex
\section{DISCUSSION}
\label{sec:discussion}

The results in Sections \ref{sec:exp} and \ref{sec:exp_usb} show that the insertion performance improves when feedback from the laser scanner is considered by the assembling arm. This is especially noticeable in the needle threading task where the tolerance is very low. 
Despite the positive results, some aspects deserve attention. In Table \ref{tab:SR_needle}, three insertions with laser scanner were unsuccessful in the first attempt. A possible explanation is the presence of outliers in the point cloud, that affect registration. In our experience, rescanning and trying again (after the failure) can help. However, detecting failure might require a human operator to monitor the scene which may hinder the application in high-volume scenarios. Detecting failure and fully automating the solution was beyond the scope of this paper.

\begin{figure}
	\centering
	\includegraphics[width=1.0\columnwidth]{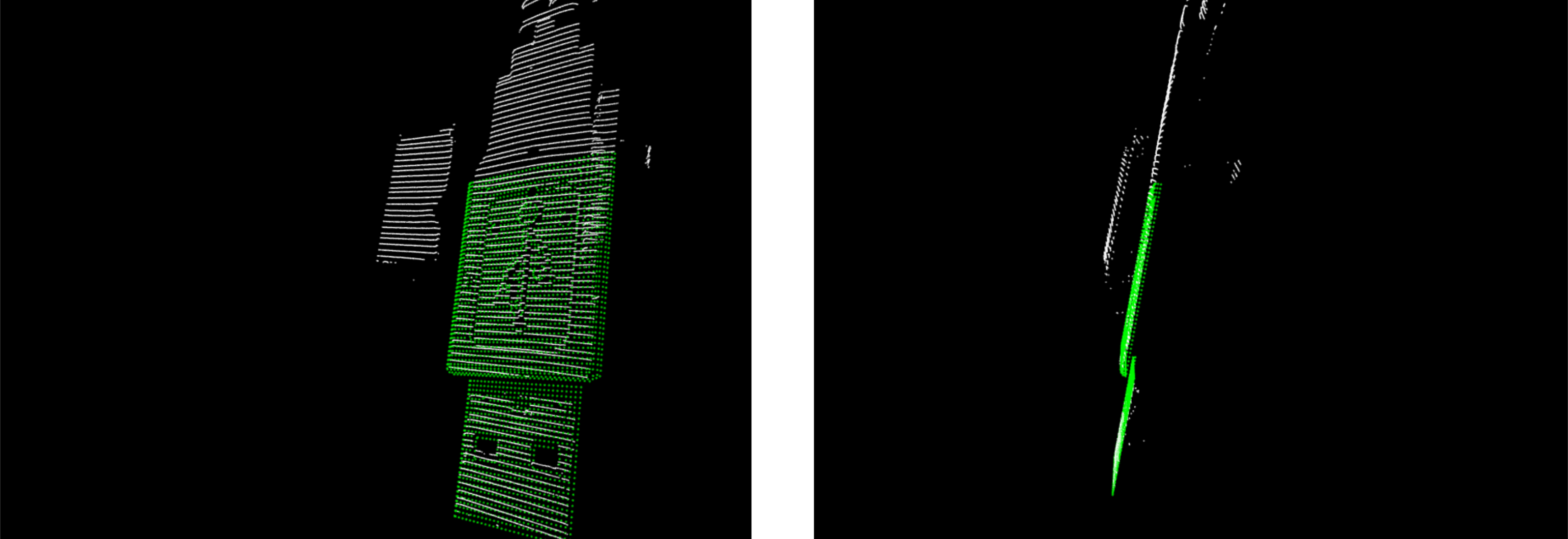}
	\caption{Front and side view of the USB plug registration. \textbf{White:} Scanned point cloud $\mathcal{P}_{scan}$. \textbf{Green:} Point cloud $\mathcal{P}_{ref}$, obtained from the CAD model (here aligned with $\mathcal{P}_{scan}$ after Algorithm \ref{algo:pose_estimation}). Since the plug was scanned from the front, we relied on a subpart of the whole CAD model. }
	\label{fig:USB_pc}
\end{figure}

In the experimental evaluations only position correction was required but some tasks may require orientation as well. 
Correcting for both position and orientation simultaneously can be difficult since orientation motions may require too large joint displacements, leading to significant (at this scale) pose errors.
One possible way to use our approach in those scenarios is to split the insertion into two phases, where the position and orientation errors are corrected sequentially.



Finally, since the scanning arm is itself a collaborative robot, it is recommended to minimize the amplitude of joint motions during scans, so as to mitigate kinematic errors as much as possible (which could lead to inaccurate point clouds).
%
%
One might also be tempted to mount the sensor on the assembling arm (to use just one robot) but since the required motion between scan and insertion could be large, there is a high chance that too much kinematic error is accumulated and the task is compromised.



%% file: sections/conclusion.tex
\section{CONCLUSIONS}
\label{sec:conclusion}

We proposed a solution that enables collaborative robots to perform high precision tasks. We have shown that, using our approach, a Panda robot could insert a $150\mu m$ thread into a $300\mu m$ needle hole with a success rate close to 100\%. The approach consists of a dual-arm system where one arm controls the motion of a high-resolution laser scanner while the other performs the insertion. It relies on the registration of scanned point clouds to find poses of objects of interest and plan an insertion trajectory that corrects initially imprecise ones.  It is particularly well suited to low-volume, high-mixture manufacturing scenarios, where collaborative robots are meant to excel regardless of object size.

In future work we plan to extend the approach to consider the end-effector and object orientations, and perform insertions in optimal ways by relying on model predictive control formulations on Riemannian manifolds \cite{Calinon20}. The possibility to use other benchmarking protocols 
(e.g. \cite{Chatzilygeroudis20}), additional metrics such as completion time \cite{Kimble20} and to lower the total cost of the solution will also be studied.